\documentclass[10pt,twocolumn,letterpaper]{article}

\usepackage{cvpr}
\usepackage{times}
\usepackage{epsfig}
\usepackage{graphicx}
\usepackage{amsmath}
\usepackage{amssymb}
\usepackage{tabularx}
\usepackage{color}
\usepackage{multirow}
\usepackage{authblk}

\usepackage[breaklinks=true,bookmarks=false]{hyperref}

\cvprfinalcopy % *** Uncomment this line for the final submission

\def\cvprPaperID{****} % *** Enter the CVPR Paper ID here

\begin{document}

%%%%%%%%% TITLE
\title{Learning to Act Properly: Predicting and Explaining Affordances from 
Images}

\author{Ching-Yao Chuang$^1$,\enspace Jiaman Li$^1$,\enspace Antonio Torralba$^2$,\enspace Sanja Fidler$^1$\\
\vspace{-3mm}
$^1$University of Toronto\enspace$^2$Massachusetts Institute of Technology\\
{\tt\small \{cychuang, ljm, fidler\}@cs.toronto.edu\enspace torralba@mit.edu}
}

%\author[1]{Ching-Yao Chuang}
%\author[1]{Jiaman Li}
%\author[2]{Antonio Torralba}
%\author[1]{Sanja Fidler}
%\affil[1]{University of Toronto}
%\affil[2]{Massachusetts Institute of Technology}

\pagestyle{empty} 

\maketitle
\thispagestyle{empty}

%%%%%%%%% ABSTRACT
\begin{abstract}
We address the problem of affordance reasoning in diverse scenes that appear in the real world. Affordances relate the agent's actions to their effects when taken on the surrounding objects. In our work, we take the egocentric view of the scene, and aim to reason about action-object affordances that respect both the physical world as well as the social norms imposed by the society. We also aim to teach artificial agents why some actions should not be taken in certain situations, and what would likely happen if these actions would be taken. We collect a new dataset that builds upon ADE20k~\cite{zhou12017ade}, referred to as ADE-Affordance, which contains annotations enabling such rich visual reasoning.  We propose  a model that %builds a graph based on object instances in an image and 
exploits Graph Neural Networks to propagate contextual information from the scene in order to perform detailed affordance reasoning about each object. Our model is showcased through various ablation studies, pointing to successes and challenges in this complex task.

\iffalse
% OLD VERSION
   We address the problem of affordance reasoning in real world. An affordance is an action that is suggested or somehow implied to an agent capable of performing that action by an object or situation in the agent’s immediate environment. For an intelligent robot acting in real world, it's essential for it to correctly reason the affordance in order to prevent negative or even perilous consequence. Given an image, the task is to predict the affordance for each action in egocentric fashion. Various circumstances will happen in real world, correctly predicting the affordances thus requiring model to perform reasoning at different levels. We propose a model based on Graph Neural Networks that allows us to efficiently analyze the relationships between the objects in image and jointly predict and explain affordances. In order to test our proposed method, we introduce the ADE20K-Affordance dataset consisting of 10000 densely annotated images. We evaluate various tasks and our method consistently outperforms other baselines.
   \fi
\end{abstract}

%%%%%%%%% BODY TEXT
\section{Introduction}
\label{sec:intro}

Autonomous agents are not simply passive observers but need to be active in an environment. For example, we might want them to fetch us dinner from a restaurant, take the dog for a walk, or carry our bags at the airport. While this seems futuristic today, we may quite likely see robots mingling among us in the near future. In order to blend naturally within the society,  robots will need to act as humans would. 
Thus, these agents will need to  understand both, the affordances and constraints imposed by the 3D environment, as well as what actions are socially acceptable in a given scene. 
For example, running into physical obstacles is dangerous for an agent and thus should be avoided. Furthermore, sitting on a chair with a purse on top is not appropriate as the chair is likely reserved by someone else. Running in a classroom during a lecture is also not conventional.

Introduced by Gibson~\cite{gibson1979}, affordances define the relationship between an agent and the environment by means of the agent's perception. Thus, they relate the agent's actions to their effects when taken upon the surrounding objects. In our work, we are interested in reasoning about action-object affordances that respect both the physical world as well as the social norms imposed by the society. This is a very challenging and unexplored problem. Understanding such affordances cannot be determined by, for example, building a simple knowledge base, as they are a function of the complete dynamic scene around us. 

We adopt the egocentric view, where a given photo is considered as the current view of the agent.  Given a set of actions that the agent is aiming to take, our goal is to find all objects/places in the scene for which taking such an action would be safe and appropriate. We further aim to teach the agent why certain objects, to which these actions could normally be applied to, are considered an exception in the current scene. Moreover, we want the agent to also understand the most likely consequence to occur if this action would in fact be taken with such an object. 

\begin{figure}[t!]
\vspace{-2mm}
\begin{center}   \includegraphics[width=\linewidth]{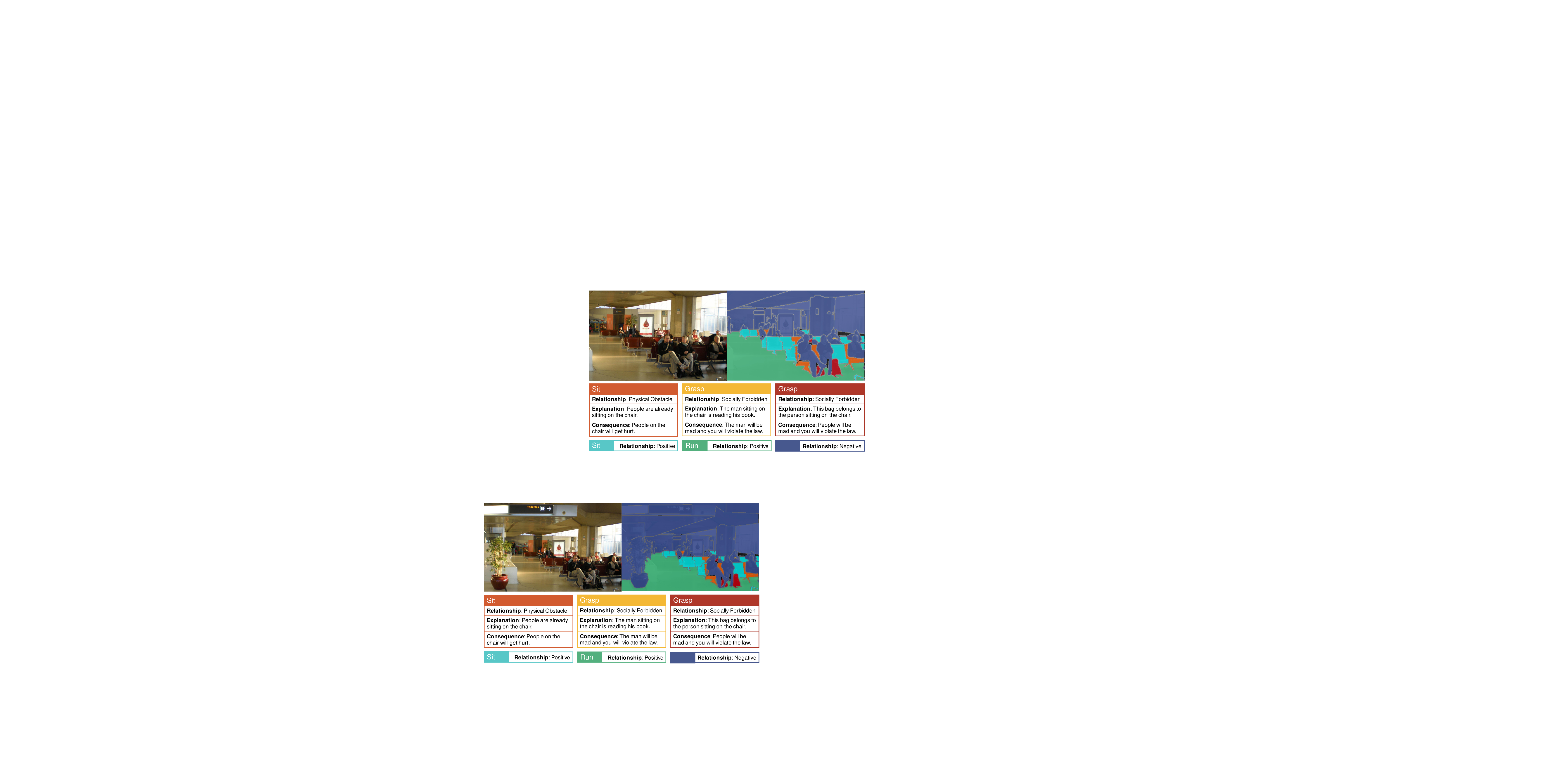}
\end{center}
\vspace{-5mm}
\caption{Example of Affordance in Real World}
\label{figure1}
\vspace{-3mm}
\end{figure}

Since no such information exists to date, we collect a new dataset building upon ADE20k~\cite{zhou12017ade} containing 20 thousand images taken in various types of scenes. In particular, we first crowdsource typical affordances for action-object pairs, thus giving us a list of all object classes to which a particular action can normally be applied to. We then mark all objects belonging to these classes in ADE20k imagery, and ask annotators to mark exceptions, i.e., objects to which this action should not be taken in this particular circumstance. We consider multiple types of exceptions, such as ``not safe'', ``not physically possible'',  or ``socially awkward''.
For each exception, we further crowdsource an explanation (reason) as well as the most probable consequence, both taking the form of a natural language sentence.

We then design a model to generate such rich visual reasoning about an image in an automatic fashion. We exploit Graph Neural Networks (GNNs) to reason about the affordances about objects in images given actions of interest. We build a graph based on an instance level semantic segmentation map, where the nodes are the objects in the image. We encode the spatial relationship of pairs of nodes by connecting adjacent objects with edges. The GNN model then takes semantic feature representation of each object as its initial node representation, and iteratively updates its hidden vectors by propagating messages among the neighbors in the graph. This allows us to capture contextual relationships in the image in an efficient way. Our model then outputs an action-object affordance for each node, as well as an explanation and consequence using an RNN language model. 

We showcase our model through various ablation studies, leading to insights about what type of information is important to perform such involved affordance reasoning. Our dataset is available at \url{http://www.cs.utoronto.ca/~cychuang/learning2act/}. 

\section{Related Work}

We review works most related to ours, focusing on affordances, visual reasoning, and captioning.

\vspace{-3mm}
\paragraph{Affordance Reasoning.} A number of works investigated learning and inferring object-action affordances~\cite{Grabner11,Shu16,ChaoWMD15,Kjellstrom10,Jiang13,zhu2014reasoning}.  Most approaches are designed to learn typical action-object affordances from visual or textual information.  In~\cite{Grabner11,Jiang13}, the authors learn visual cues that support a particular action, such as sit, by imagining an actor interacting with the object. ~\cite{Kjellstrom10} learns object-action affordances by watching humans interacting with objects, while~\cite{Shu16} infers social interactions by observing humans in social contexts. In~\cite{ChaoWMD15,Yao13,zhu2014reasoning}, the authors learn affordance relations between actions and objects by either crowd-sourcing, mining textual information, or  parsing images. However, most of these approaches treat affordances as static, i.e., applicable to all situations, and mainly focus on learning what object attributes imply certain affordances. In our work, we specifically reason about scene-dependent exceptions, both physical and social, and further aim to explain them. Recently,~\cite{alayrac2017joint} aimed at inferring object states, such as \emph{open} or \emph{close}, which is an important cue for affordance reasoning. Object state depends on the visual observation of the object alone, while in our work we are interested in going beyond this by reasoning about affordances in the scene as a whole. 

\vspace{-3mm}
\paragraph{Visual Description Generation.} 
Describing visual content is a fundamental problem in artificial intelligence that connects computer vision and natural language processing. Image captioning approaches~\cite{karpathy2015cap, vinyals2015sat, xu2015saat, chen2017sat,DaiICCV17} generate a natural language sentences to describe an entire image. %They typically employ a Convolutional Neural Networks (CNNs) for image encoding, then decoding a caption with a Recurrent Neural Networks (RNNs). 
In~\cite{densecap}, the authors aim at both, generating region detectors and describing them in language, sharing some similarity with our effort here. 
Recently,~\cite{barratt2017interpnet,Hendricks16} proposed a framework to generate explanations for their classification decisions. %This work is somewhat related to our efforts here. 
However, different from these works, we generate an explanation per object and exploit dependencies between objects in an image to perform this reasoning.~\cite{Vicol17} used  graphs to generate explanations for human interactions. 

\vspace{-3mm}
\paragraph{Graph Neural Networks.} Several approaches apply deep neural networks to graph structured data. One direction is to apply CNNs to graphs in the spectral domain by utilizing the graph Laplacian~\cite{bruna2014graph, defferrard2014graph, kipf2017graph}. ~\cite{duvenaud2015graph} designed a hash function such that CNN can be applied to graphs. We here adopt the Gated Graph Neural Networks~\cite{li2015gated} that define a GRU-like propagation model on each node of the graph. 

We are not the first to use GNNs to model dependencies in vision tasks. In~\cite{SituationsICCV17},  the authors exploit a GGNN~\cite{li2015gated} to predict situations from images in the form of an action and role-noun assignments. ~\cite{3dggnnICCV17} used GGNN to perform semantic segmentation in RGB-D data. ~\cite{liang2016semantic} defines graph LSTM over the nested superpixels in order to perform scene parsing. In our work, we exploit GGNN to model dependencies among objects in order to produce affordances and their explanations.

\vspace{-1mm}
\section{ADE-Affordance Dataset}
Since we are not aware of existing datasets tackling visual reasoning about action affordances, we collected a new dataset which we refer to as the ADE-Affordance dataset. We build our annotations on top of the ADE20K~\cite{zhou12017ade}. ADE20k contains images from a wide variety of scene types, ranging from indoor scenes such as \emph{airport terminal} or \emph{living room}, to outdoor scenes such as \emph{street scene} or \emph{zoo}. It covers altogether 900 scenes, and is a good representative of the diverse world we live in.  Furthermore, ADE20k has been densely annotated with object instance masks, forming likely one of the most comprehensive datasets to date. 

\subsection{Dataset Collection}

We divide our annotation effort into three stages: {\bf 1)} creating a knowledge base (KB) of typical action-object affordance pairs, {\bf 2)} marking KB objects in images with exception types (object class is in KB, however in this particular scene the action cannot or should not be applied to the object instance), and {\bf 3)} collecting explanations and most probable consequences for such special objects. We describe our full collection pipeline in this section. 

\begin{figure}[t!]
\vspace{-2mm}
\begin{center}   
\includegraphics[width=1\linewidth]{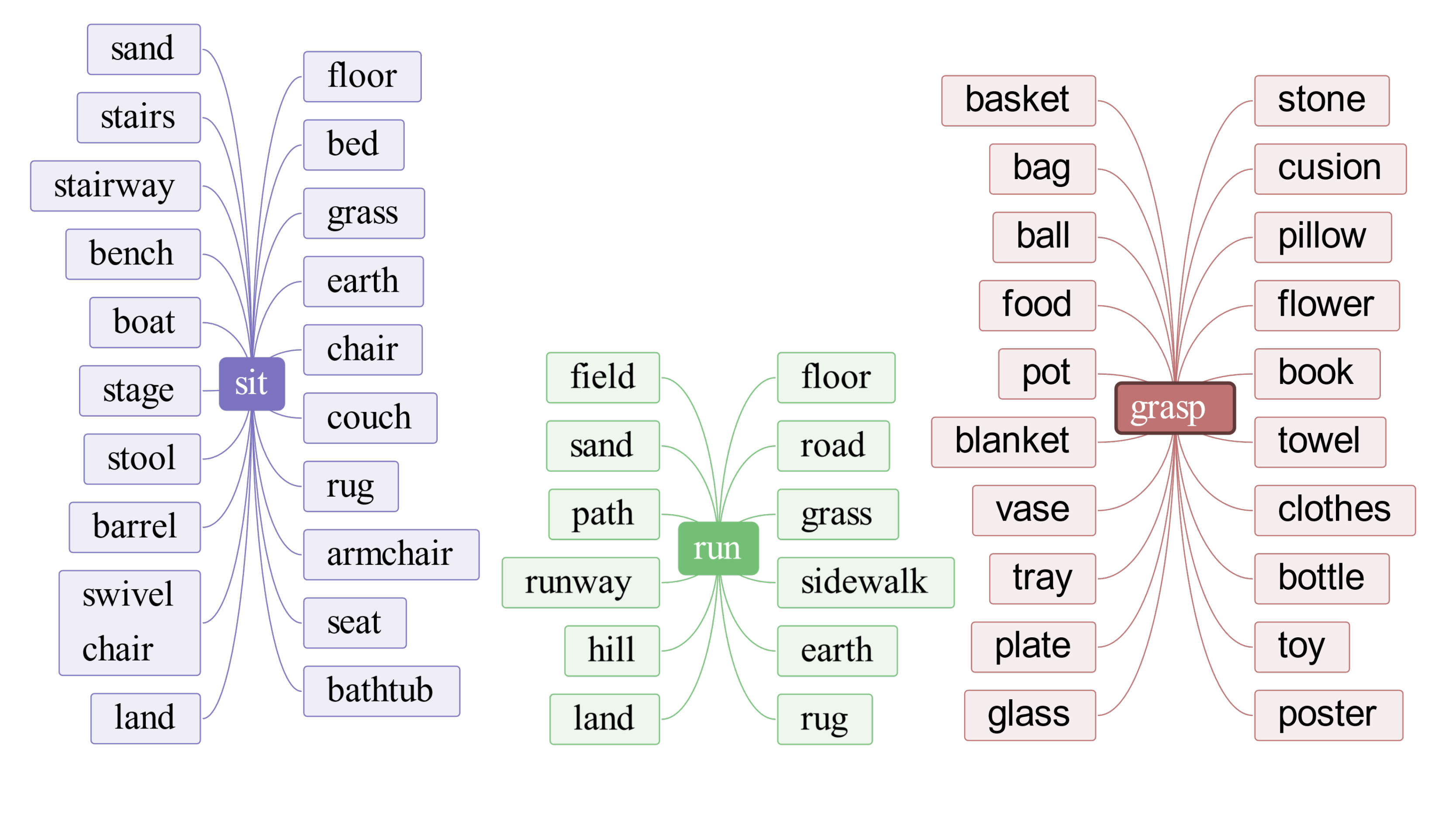}
\end{center}
\vspace{-5mm}
\caption{Knowledge Base of action-object affordances}
\label{kb}
\vspace{-3mm}
\end{figure}

\vspace{-3mm}
\paragraph{Collecting the Affordance Knowledge Base.}
While objects in ADE20k were annotated using an open vocabulary, the authors also provide a list of 150 classes which cover most of the objects in the dataset. In the first step of our data collection, we asked AMT workers to list five most typical actions for each of the 150 object classes. We asked three workers per object class. We then chose three actions which we found to be elementary for an agent, as well as interesting: \emph{sit}, \emph{run}, \emph{grasp}. For each action, we then compiled a list of all object classes from the AMT annotations. For example, \emph{bottle} is linked to action \emph{grasp}, and \emph{floor} is linked to \emph{run} and \emph{sit}, among others.
We further added a few more objects to the list if we felt that it was not sufficiently comprehensive. This list (compiled for each of the three actions)  forms our affordance knowledge base. Note that this KB is not complete as it may miss objects outside the 150 classes. We augment our KB with such information in the following stages of data collection. We show our KB in Fig.~\ref{kb}. We then select $10,000$ images from ADE that contains most object classes from our KB, and build our dataset on top of this subset. 

\vspace{-3mm}
\paragraph{Annotating Exceptions in Images.} We first carefully considered the types of exceptions that prevent actions to be taken with certain objects. In particular, we chose seven different action-object relationships, one positive (the action can be taken), one firmly negative (the action can never be taken with this object class -- this comes from our KB), and five exception types (normally one could take an action with this object class, however not with this particular object instance). Table~\ref{table:relationship} lists our chosen set of relationships.

\begin{figure*}[h]
\vspace{-2mm}
\begin{center}
\resizebox{2\columnwidth}{!}{
\includegraphics[width=1\linewidth]{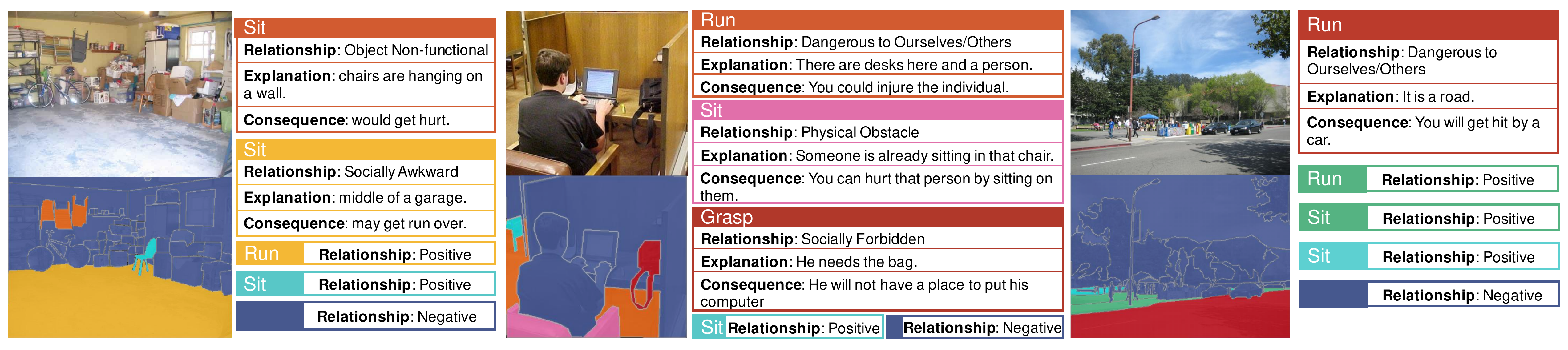}
}
\end{center}
\vspace{-4mm}
\caption{Examples of annotations from our ADE-Affordance dataset}
\label{dataexp}
\vspace{-1mm}
\end{figure*}
We then designed an annotation webpage which shows the original image, as well as the image containing object instance  masks.  The data for each action was collected separately to prevent bias. For each action, we marked all object instances from its KB in green, indicating a potential positive affordance, and the negative objects in blue (negative affordance). We asked the annotators to help a robot understand which objects form the exception.  We wrote detailed visual instructions explaining what we want, with additional examples further illustrating the task.

The annotator was requested to click on the object and was required to select 1 out of 5 possible exception classes. The annotator was also able to add positive affordances, by clicking on any blue (negative) object, if they felt that the object could in fact afford an action. We additionally asked the annotators to select any green (positive) objects and add exceptions to them if they felt that this type of object typically affords the action, but forms an exception in this case. These are the objects that have been missed by our KB. 

At the end of this data collection, we have all positive action-object instances marked for each image, and for each of the negative action-object instances we additionally have the exception type, helping the agent to understand why the action should not be taken. 

\vspace{-3mm}
\paragraph{Adding Explanations and Consequences.} The previous stage of data collection yields a broad exception category for certain objects. Here, we are interested in obtaining a more detailed understanding of the exception. In particular, we now show the annotator only the objects marked as exceptions, and ask them to write an explanation as well as a likely consequence for each of them. To illustrate, consider an action \emph{grasp} and an image that shows a woman holding her purse. The exception type for the purse would be marked as \emph{socially forbidden}. Here, a plausible explanation could be: \emph{One cannot take other people's property}. The likely consequence if one would still attempt to grasp the purse would be: \emph{One could go to jail}. We advised the annotators to write sentences in third person. 

Note that due to the somewhat subjective flavor of the task, we collected annotations from three workers per image for our \emph{test} subset of the dataset, allowing more robust evaluation of performance. For the \emph{train} and \emph{val} subsets, we only collected annotation from one worker. We show a few examples of complete annotations in Fig.~\ref{dataexp}.

\begin{figure}[t!]
\vspace{-4mm}
\begin{center}   \includegraphics[width=1\linewidth]{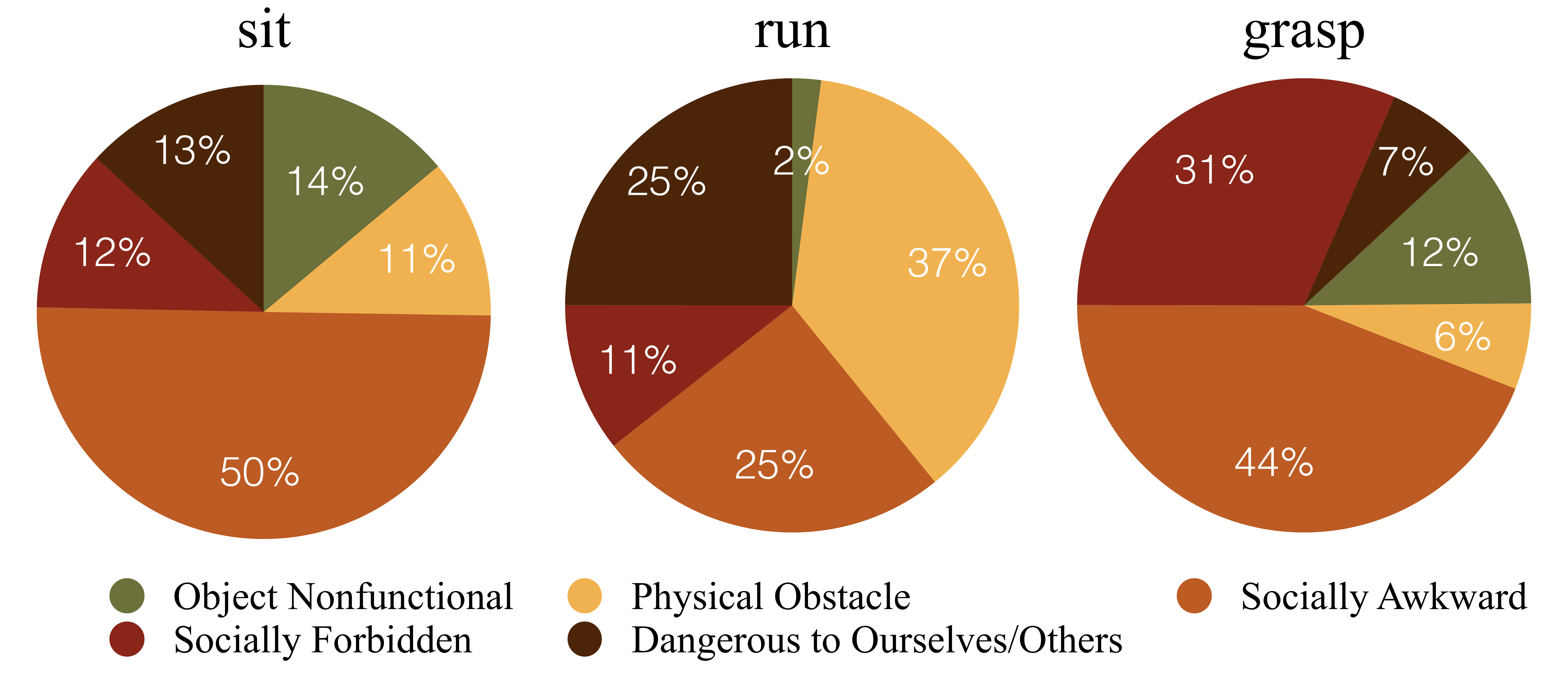}
\end{center}
\vspace{-3.5mm}
   \caption{Exception distribution for each action}
\label{exception_type}
\vspace{-2.0mm}
\end{figure}

\begin{figure}[t!]
\vspace{-1mm}
\begin{center}
\includegraphics[width=1\linewidth]{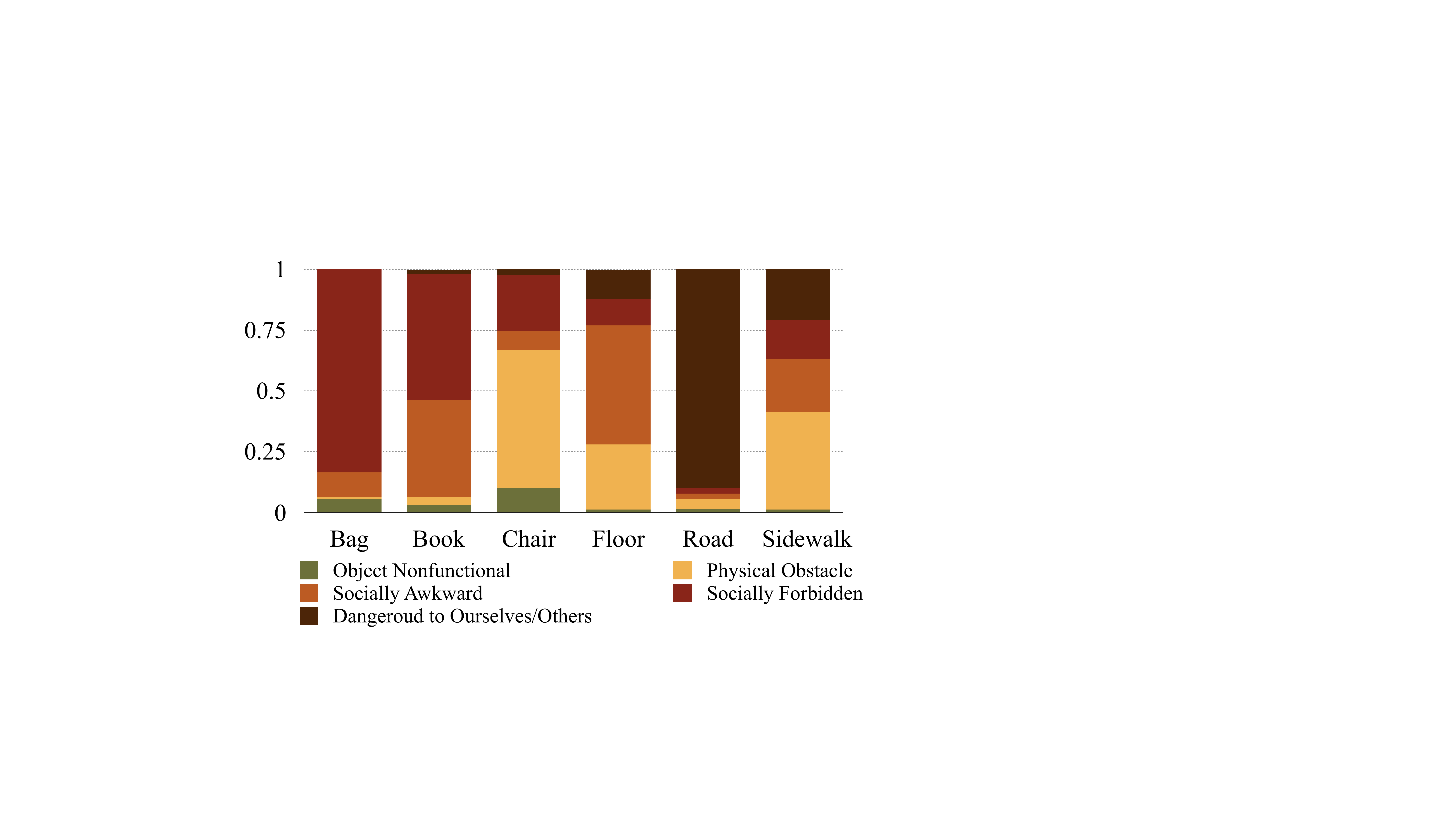}
\end{center}
\vspace{-3.5mm}
   \caption{Exception distribution for selected objects}
\label{obj_exception}
\vspace{-0.0mm}
\end{figure}

\subsection{Dataset Analysis}

\begin{table}
\begin{center}
%\centering
\resizebox{\columnwidth}{!}{
\begin{tabular}{@{}|l|c|c@{}} 
%\begin{tabular}{|c|c|} 
 \hline
 Relationship &  Description \\ \hline\hline
 Positive & We can take this action to the object\\ \hline
 Firmly Negative & We can never take this action to the object\\ \hline
 Object Non-functional & The object lose some of its original function \\ \hline
 Physical Obstacle & Physical obstacles prevent you to take the action.  \\ \hline
 Socially Awkward & It is not proper to take the action  \\ \hline
 Socially Forbidden & It is strictly forbidden to take the action   \\ \hline
 Dangerous to ourselves/others & Taking the action would harm ourselves or others  \\ \hline

\end{tabular}
}
\end{center}
\vspace{-2.5mm}
\caption{\small Action-Object Relationship Categories}
\label{table:relationship}
\vspace{-2mm}
\end{table}

\begin{table}[t!]
\begin{center}
%\centering
\resizebox{1\columnwidth}{!}{
\begin{tabular}{@{}lcccc@{}} 
%\begin{tabular}{|c|c|c|c|} 
 \hline
  &  Sit & Run & Grasp\\ \hline\hline
 Images with Exception (\%) & 52.2 & 77.5 & 22.7 \\ \hline
 Objects with Exc. (\%) wrt KB classes & 23.2 & 41.7 & 13.0 \\ \hline
 Objects with Exc. (\%) wrt all objects & 4.33 & 4.35 & 1.96 \\ \hline
\end{tabular}
}
\end{center}
\vspace{-2.5mm}
\caption{\small Statistics of exception annotations in our dataset}
\label{table:exception-partition}
\vspace{-1mm}
\end{table}

\begin{table}[t!]
\vspace{-1mm}
\begin{center}
%\centering
\resizebox{0.95\columnwidth}{!}{
\begin{tabular}{@{}lcccccc@{}} 
%\begin{tabular}{|c|c|c|c|} 
 \hline
  &  & \#Sent. & Vocab. & Avg Len. & \#Sent./W. \\ \hline\hline
 \multirow{2}{*}{Sit} & Explanation & 14221 & 3768 & 7.75 & 69.5 \\ 
 & Consequence & 14221 & 3636 & 7.76 & 75.8 \\ \hline
 \multirow{2}{*}{Run} & Explanation & 13914 & 2182 & 6.89 & 108.6\\ 
 & Consequence & 13914 & 2098 & 7.63 & 132.4 \\ \hline
\multirow{2}{*}{Grasp} & Explanation & 9785 & 2341 & 7.81 & 84.1 \\ 
& Consequence & 9785 & 2178 & 7.88 & 63.6 \\ \hline
\end{tabular}
}
\end{center}
\vspace{-1.5mm}
\caption{\small Statistics of explanation and consequence sentences. \emph{Voca.} stands for vocabulary. \emph{Sent.} stands for sentences. \emph{W.} stands for words. \emph{Len.} stands for length of sentences.}
\label{table:text-statistic}
\vspace{-2mm}
\end{table}

%In Table~\ref{table:exception-partition} 
We analyze our dataset through various statistics. In Table~\ref{table:exception-partition}, we show the proportion of images that contain at least one object with exception for each of the actions. For \emph{sit} and \emph{run}, exceptions exist in more than 50\% of images, showing that special conditions are common in real world scenes. We further computes the percentage of objects belonging to classes in our KB that are assigned exceptions in images, as well as the percentage of exceptions with respect to all objects in our ADE-Affordance dataset.  

%Table~\ref{table:exception-partition} show the partition of exceptions. For sit and run, exceptions exist in more than fifty percents of images, showing that special condition is common in real world scenes. Object Partition without negative objects indicates the proportion of exception in all the objects without objects you can never take the action to it and object partition with negative objects includes them.

Fig~\ref{exception_type} shows the distribution of different exceptions classes for the three actions. We can see that the ratio of \emph{Socially Awkward} is higher for \emph{sit} and \emph{grasp} while \emph{run} has more \emph{Physical Obstacle} exceptions, indicating that we encounter different situations while taking different actions. Fig~\ref{obj_exception} shows the distribution of different exceptions classes for a few common objects in our dataset. We can see very specific exception ``signatures'' for different objects.

Table~\ref{table:text-statistic} shows  statistics of explanation/consequence sentences for the three actions. The first two columns show the number of sentences and non-stemmed vocabulary size, respectively. %The third column is the average length of the sentences. The forth column is the average number of sentences per word. 
We can observe that the sentences for the action \emph{sit} are more complex than for \emph{run} and \emph{grasp}. The forth column is the average number of sentences per word.

We split our dataset into 8,011 training images, 1000 images in the validation set, and 1000 images in the testing set, by balancing exception classes across the splits.

\section{Affordance Reasoning with Graph Neural Models}
\label{sec:method}

\begin{figure*}[h]
\vspace{-2mm}
\begin{center}   
\includegraphics[width=1\linewidth]{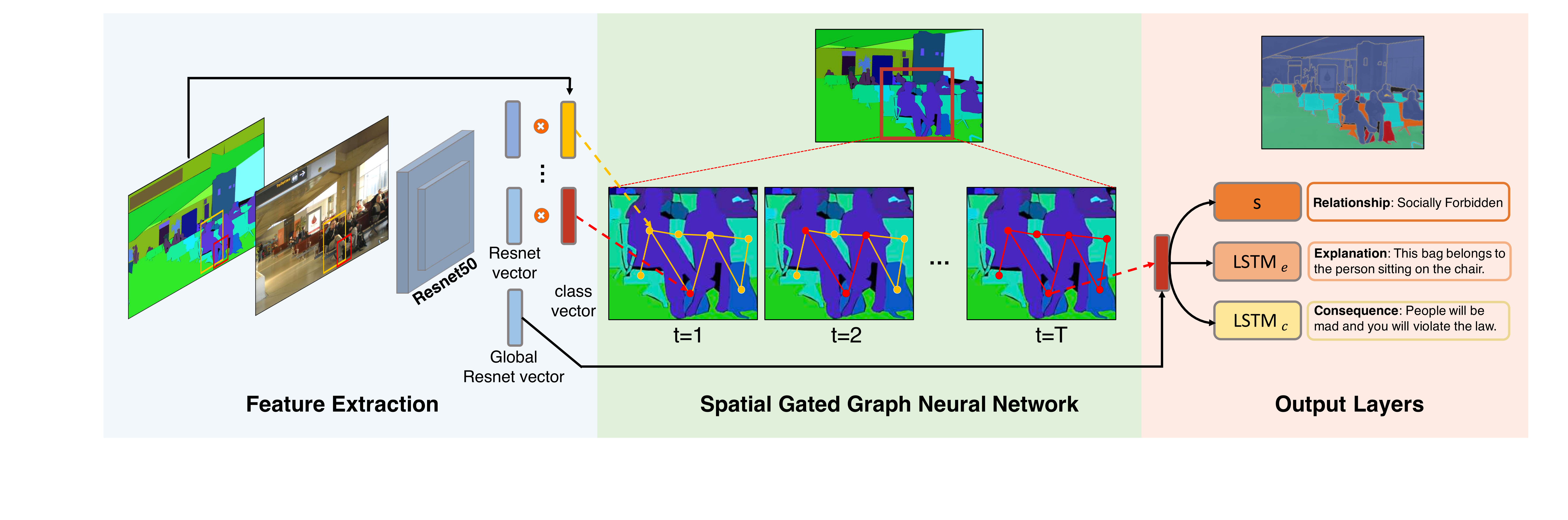}
\end{center}
\vspace{-5mm}
\caption{Model Architecture}
\label{model_pipeline}
\vspace{-4mm}
\end{figure*}

In this section, we propose a model to perform visual reasoning about action-object affordances from images. We first formalize the task, and introduce our graph-based neural model in the following subsection.

\subsection{Problem Definition}
We define the set of  $N$ actions as $\mathcal{A} = {\{\mathbf{a}^i\}}_{i=1}^N$, where $N=3$ in our case. Given an image $I$, let $M$ be the number of object instances $\mathbf{O} = {\{\mathbf{o}^j\}}_{j=1}^M$ that compose the scene. These ``objects'' include discrete classes such as \emph{table} and \emph{chair}, as well as ``stuff'' classes such as \emph{floor} and \emph{road}. Our task is to predict and explain affordances $\mathbf{A}(\mathbf{a}^i, \mathbf{o}^j)$ for each action-instance pair. In particular,  $\mathbf{A}(\mathbf{a}^i, \mathbf{o}^j)$ includes the following information: 

\vspace{2mm}
\noindent\textbf{1. Relationship}: For each instance $\mathbf{o}^j$, we categorize its relationship with action $\mathbf{a}^i$ as ${\mathbf{r}}_{j}^i \in [\mathcal{R}_{pos}, \mathcal{R}_{neg}, \mathcal{R}_{exc}]$,  indicating the scene dependent affordance of $\mathbf{a}^i$ and $\mathbf{o}^j$. Here, $\mathcal{R}_{pos}$ indicates that action $\mathbf{a}^i$ can be taken with object $\mathbf{o}^j$, while $\mathcal{R}_{neg}$ indicates that action $\mathbf{a}^i$ cannot be taken with $\mathbf{o}^j$ under any condition (such as \emph{grasp} with \emph{sky}). $\mathcal{R}_{exc}$ contains several types of exceptions indicating the special circumstances due to which the action cannot be taken.

\vspace{2mm}
\noindent\textbf{2. Explanation}: For each $(\mathbf{a}^i, \mathbf{o}^j)$ pair for which ${\mathbf{r}}_{j}^i \in \mathcal{R}_{exc}$, we aim to explain the reason why the action cannot be taken. This explanation is a natural language sentence.

\vspace{2mm}
\noindent\textbf{3. Consequence}: For each $(\mathbf{a}^i, \mathbf{o}^j)$ pair for which ${\mathbf{r}}_{j}^i \in \mathcal{R}_{exc}$, foresee the (negative) consequence if action $\mathbf{a}^i$ was to be taken. This consequence is a natural language sentence.

These three entities serve as the essential information to understand the dynamics in the real world and help us reason about the proper way to interact with the environment.

\vspace{-0mm}
\subsection{Our Model}
Action-object affordance depends on the contextual information in the scene. For example, we can not sit on the chair occupied by another person, and we should not cross the road with cars rushing upon it. Reasoning about the affordances thus requires both, understanding the semantic classes of all objects in the scene, as well as their spatial relations. We propose to model these dependencies via a graph $G=(\mathcal{V}, \mathcal{E})$. The nodes $v \in \mathcal{V}$ in our graph indicate objects in the image, while the edges $e \in \mathcal{E}$ encode the spatial relations between adjacent objects. We define two objects to be adjacent if their boundaries touch. 

\vspace{-3mm}
\paragraph{Background.}
Gated Graph Neural Network (GGNN) is a neural network that propagates information in a graph, and predicts node or graph-level output. Each node of a GGNN is represented with a hidden vector that is updated in a recurrent fashion. At each time step, the hidden state of a node is updated based on its previous state and messages received from the neighbors. After $T$ propagation steps, the hidden state at each node is used to predict the output.

\vspace{-3.0mm}
\paragraph{Spatial GGNN for Affordance Reasoning.}
We adopt the GGNN framework to predict and explain affordances. We refer to our method as Spatial GGNN since our graph captures spatial relations between objects to encode context. 

To initialize the hidden state for node $v$, we combine the object class information and an image feature vector: % pre-trained on image classification dataset. 
%We initialize the hidden state for each node $v$ as
\begin{eqnarray} \small
&&h^{0}_{v} = g(W_{c}\hat{c}) \odot g(W_{f}\phi(\mathbf{o}^v)))
\end{eqnarray}
where $\hat{c} \in \{0,1\}^{|\mathcal{C}|}$ corresponds to the one-hot encoding of the object class and $\phi(\mathbf{o}^v)$ represents the feature vector. In our experiments, we either use the ground-truth class or the predicted class using an instance segmentation network~\cite{qi2017fcis}. We obtain $\phi(\mathbf{o}^v)$ by cropping the image patch within the box defined by the object's mask, and extracting features from the last hidden layer of Resnet-50~\cite{he2015resnet} features. Here, $W_{c}$ and $W_{f}$ map the object class and the image feature vector to the space of the hidden representations. Note that $\odot$ corresponds to element-wise multiplication, and $g(\cdot)$ is the non-linear function ReLU.

At time step $t$, the node's incoming information is determined by the hidden state of its neighbors $\{v'\in\mathcal N(v)\}$:
\vspace{-5mm}
\begin{eqnarray} \small
&&x^{t}_{v} = \sum_{v'\in\mathcal N(v)} W_{p}{h_{v'}^{t-1}} + b_{p}
\end{eqnarray}
The linear layer $W_{p}$ and bias $b_{p}$ are shared across all nodes.

After aggregating the information, the hidden state of the node is updated through a gating mechanism similar to the Gated Recurrent Unit (GRU) as follows~\cite{li2015gated}:
\vspace{-2mm}
\begin{eqnarray} \small
&&z^{t}_{v} = \sigma(W_{z}x^{t}_{v} + U_{z}h^{t-1}_{v} + b_{z}), \nonumber\\
&&r^{t}_{v} = \sigma(W_{r}x^{t}_{v} + U_{r}h^{t-1}_{v} + b_{r}), \nonumber\\ 
&&\hat{h}^{t}_{v} = \mathrm{tanh}(W_{h}x^{t}_{v} + U_{h}(r^{t}_{v} \odot h^{t-1}_{v}) + b_{h}), \nonumber\\
&&h^{t}_{v} = (1-z^{t}_{v}) \odot h^{t-1}_{v} + z^{t}_{v} \odot \hat{h}^{t}_{v}
\end{eqnarray}

\vspace{-1mm}
While maintaining its own memory, each node can extract useful information from incoming messages.

\vspace{-3mm}
\paragraph{Output.} After $T$ propagation steps, we extract node-level hidden states $h^{T}_{v}$. Before feeding to the output layer, we combine each $h^{T}_{v}$ with the first hidden state and a global feature derived from the image. We compute the global feature $\phi(I)$ by feeding the full image into Resnet-50. The feature fed into the output layer is then computed as follows:
\vspace{-1mm}
\begin{eqnarray} \small
&&h^{o}_{v} = g(W_{ho}[h^{T}_{v}, h^{0}_{v}, \phi(I)])
\end{eqnarray}
To predict relationship, we use two FC layers $s$ and softmax:
\vspace{-4mm}
\begin{eqnarray} \small
&&p_{s} = \mathrm{softmax}(s(h^{o}_{v}))
\end{eqnarray}
For each $(\mathbf{a}^i, \mathbf{o}^j)$ pair whose predicted relationship is in $\mathcal{S}_{neg}$, we further use the standard captioning RNN architecture to generate an explanation $y_{e}$ and a consequence $y_{c}$:
\begin{eqnarray} \small
&&y_{e} = \mathrm{LSTM}_{e}(h^{o}_{v}) \nonumber\\
&&y_{c} = \mathrm{LSTM}_{c}(h^{o}_{v})
\end{eqnarray}
%Here, we use the LSTM to generate explanations and consequences. 
Note that the networks for predicting the outputs are shared across all nodes.

\vspace{-3mm}
\paragraph{Learning.}

We use the cross-entropy loss function for relationship prediction, as well as for predicting each word in the explanation/consequence. The GGNN is then trained with the back-propagation through time (BPTT) algorithm.
\section{Experiments} 
\label{sec:results}

In this section, we first provide implementation details, and then perform an extensive evaluation of our approach. 

%\subsection{Implementation Details}
\vspace{-3mm}
\paragraph{Model Training.}
%We implement the proposed models in Tensorflow~\cite{tensorflow}. 
To train our model we use the graph derived from the ground truth segmentation map.  We set the number of propagation steps in all GGNN networks to $T=3$ except for the ablation study. For relationship prediction, we train the network using Adam~\cite{kingmaba2015adam} with mini-batches of 128 samples. Initial learning rate is $1 \times 10^{-3}$ and we decay it after 10 epochs by a factor of 0.85. The LSTM for explanation and consequence generation is trained using Adam with mini-batches of 32 samples and initial learning rate of $3 \times 10^{-4}$. In all experiments, the hidden size of MLP, GGNN and LSTM is set to 128. Local and global image feature are derived using the Resnet-50~\cite{he2015resnet} pre-trained on Imagenet~\cite{imagenet}. To extract the local feature for each object, we extend the object's bounding box by a factor of $1.2$ in order to include contextual information. In all our experiments, we train different models for each task (i.e., models for relationship classification and explanation generation do not share parameters) in order to eliminate the effect across tasks during evaluation. We also show the effect of training the model jointly on all tasks later in the section.

\vspace{-3mm}
\paragraph{Data Processing.} For explanation and consequence, we prune the vocabulary by dropping words with frequency less than 2. %We also include special Begin-Of-Sentence (BOS) and End-Of-Sentence (EOS) tokens. 
We share the vocabulary for explanation and consequence within each action, however, we do not share the parameters of their LSTMs. 

%\subsection{Evaluation}

\vspace{-3mm}
\paragraph{Baselines.} Since there is no existing work on our task, we compare to a number of intelligent baselines:

{\bf KB}: We exploit the information in our knowledge base as the weakest baseline. Note that this baseline does not exploit any visual information. When presented with the class of the object instance, it simply predicts the positive/negative relationship by looking up the KB. 

{\bf Unaries}: We obtain a unary model (graph without edges) by using $T = 0$ steps of propagation in our GGNN. This model thus only uses each object's features to perform prediction, and allows us to showcase the benefit of modeling dependencies in our model. 

{\bf Chain RNN}: An unrolled RNN can be seen as a special case of a GGNN~\cite{li2017situation}, where the nodes form a chain with directed edges between them. In chain RNN  the nodes only receive information from their (left) neighbor. For this baseline, we randomly choose the order of nodes.

{\bf FC GGNN}: We simply fully connect all nodes to all other nodes without considering their spatial relationships in the image. All other settings are the same as in our model. Note that in this model, all objects contribute \emph{equally} to each of the objects, and thus the contextual information contained in the neighboring objects is not fully exploited.

\vspace{-3mm}
\paragraph{Evaluation Metrics.} Due to the complex nature of our prediction  task, we design special metrics to evaluate performance. 
For relationship prediction, we calculate two metrics: \emph{mean accuracy} (mAcc), and \emph{mean accuracy with exceptions} (mAcc-E). For mAcc, we consider all  exceptions as one class and calculate mean accuracy among 3 classes: positive, firmly negative, and the exception class. For mAcc-E, we treat different exceptions as different classes and calculate mean accuracy across all seven classes (Table~\ref{table:relationship}). When evaluating on test, we calculate each metric for each of the three GT labels, and compute the average.

For explanation and consequence generation, we use four metrics commonly used for image captioning: Bleu-4, Meteor, ROUGE, and CIDEr. On the test set, we calculate  accuracy wrt each GT sentence and take the average. %Note that the image captioning metrics are originally designed to handle multiple ground truth so we do not need further modification. 

\subsection{Results}

In all experiments, we select the model based on its accuracy on the val set, and report the performance on the test set. In order to separate the effect of doing both, instance segmentation as well as visual reasoning on the detected objects, we first evaluate our approach using ground-truth instance segmentation. 

\vspace{-3mm}
\paragraph{Relationship Prediction.}
We first consider the task of predicting the relationship between a chosen action and each object. Table~\ref{table:sinagle-relation-prediction} reports the performance. 
We use KB to set the lower bound for the task. By comparing our model to the Unaries baseline, we show that aggregating information from neighboring nodes increases the performance. The results also suggest that our graph that respects spatial relationships in images utilizes the contextual information better  than the fully-connected  and the chain graphs. 

\begin{figure*}[h]
\vspace{-2mm}
\begin{center}   
\resizebox{2\columnwidth}{!}{
\includegraphics[width=2\linewidth]{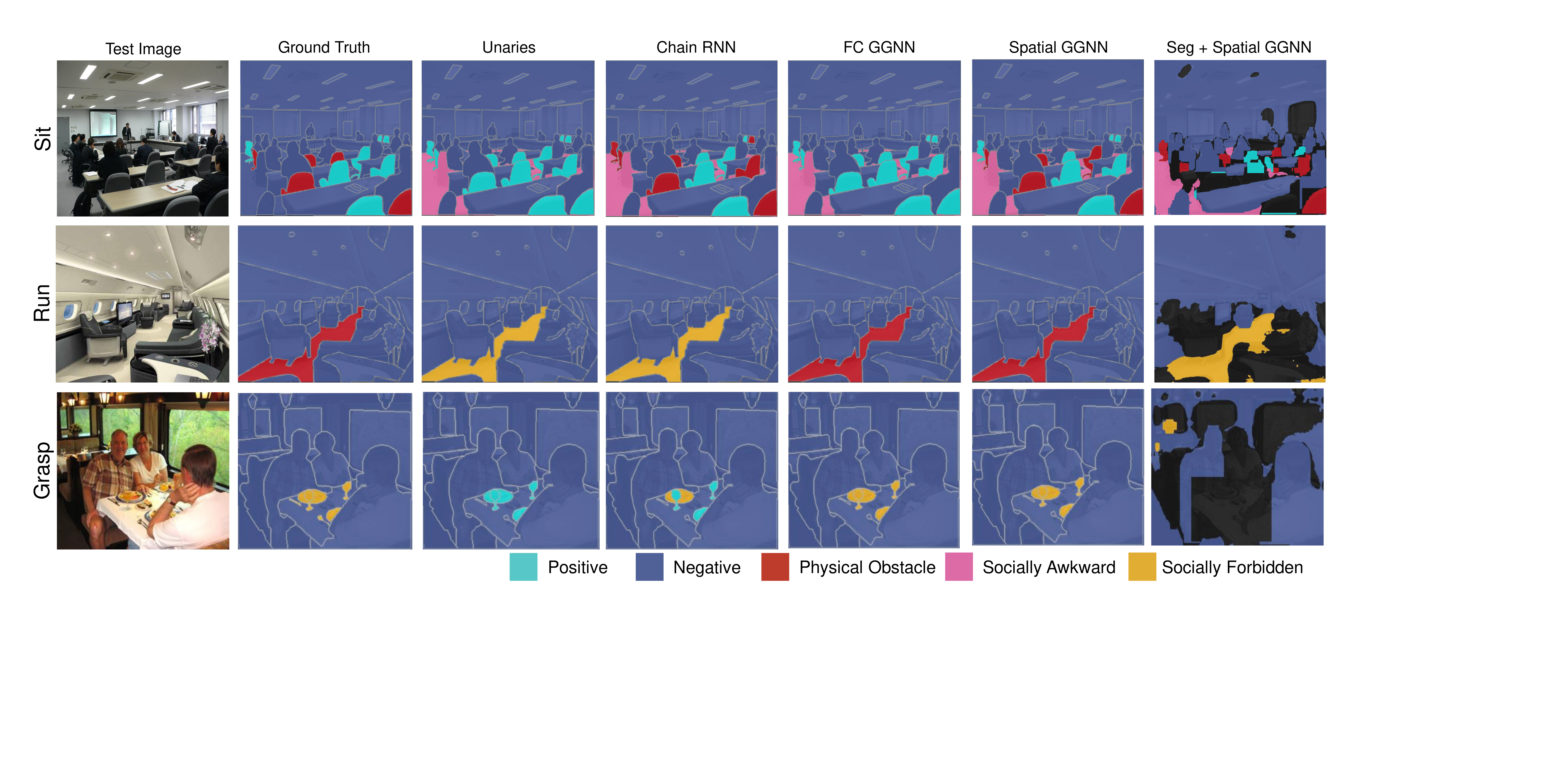}
}
\end{center}
\vspace{-3mm}
\caption{Relationship Prediction for different models. Columns 2-6 use GT segmentation, while the last row also predicts segmentation.}
\label{typical_exception}
\vspace{-4mm}
\end{figure*}

Examples in Fig.~\ref{typical_exception} show that our model is able to correctly predict exceptions by utilizing spatial information. In the first row, we can see that Unaries and FC GGNN fail to predict \emph{physical obstacle} due to the lack of spatial context. Without correct order, Chain RNN also fails to accurately detect the exception for each object.

\begin{table}[t!]
\begin{center}
%\centering
\resizebox{\columnwidth}{!}{
\begin{tabular}{@{}l|cc|cc|cc@{}}
%\begin{tabular}{|c|c|c|c|c|c|c|} 
 \hline
 \multirow{2}{*}{Method} & \multicolumn{2}{c|}{Sit} & \multicolumn{2}{c|}{Run} & \multicolumn{2}{c}{Grasp} \\ 
  &  mAcc & mAcc-E & mAcc & mAcc-E & mAcc & mAcc-E\\ \hline\hline
 KB & 0.656 & 0.281 & 0.662 & 0.284 & 0.590 & 0.253 \\ \hline
 Unaries & 0.730 & 0.428 & 0.714 & 0.434 & 0.594 & 0.289 \\ \hline
 Chain RNN & 0.727 & 0.441 & 0.720 & 0.412 & \textbf{0.610} & 0.289\\ \hline
 FC GGNN & 0.733 & 0.438 & 0.711 & 0.407 & 0.596 & 0.309\\ \hline
 Spatial GGNN & \textbf{0.745} & \textbf{0.461} & \textbf{0.730} & \textbf{0.452} & 0.604 & \textbf{0.310}\\ \hline
\end{tabular}
}
\end{center}
\vspace{-1mm}
\caption{\small Relationship Prediction using ground-truth segmentation.}
\label{table:sinagle-relation-prediction}
\vspace{-2mm}
\end{table}

\begin{table}[t!]
\vspace{-0.5mm}
\begin{center}
%\centering
\resizebox{\columnwidth}{!}{
\begin{tabular}{@{}l|cc|cc|cc@{}} 
%\begin{tabular}{|c|c|c|c|c|c|c|} 
 \hline
 \multirow{2}{*}{Method} & \multicolumn{2}{c|}{Sit} & \multicolumn{2}{c|}{Run} & \multicolumn{2}{c}{Grasp} \\ 
  &  mAcc & mAcc-E & mAcc & mAcc-E & mAcc & mAcc-E\\ \hline\hline
 KB & 0.228 & 0.098 & 0.196 & 0.084 & 0.128 & 0.055 \\ \hline
 Unaries & \textbf{0.280} & 0.156 & 0.240 & 0.190 & 0.130 & 0.059 \\ \hline
 Chain RNN & 0.272 & 0.152 & 0.241 & 0.172 & 0.130 & 0.058\\ \hline
 FC GGNN & 0.278 & 0.157 & \textbf{0.246} & 0.179 & \textbf{0.131} & 0.064\\ \hline
 Spatial GGNN & \textbf{0.280} & \textbf{0.162} & 0.238 & \textbf{0.192} & 0.130 & \textbf{0.065}\\ \hline
\end{tabular}
}
\end{center}
\vspace{-1mm}
\caption{\small Relationship Prediction with real segmentation results.}
\label{table:segmentation-relation-prediction}
\vspace{-5mm}
\end{table}

\vspace{-3mm}
\paragraph{Explanation Generation.}
We now consider generating explanations for each object with exception. The performance is shown in Table~\ref{table:explanation-generation}. GGNN structured models consistently outperform the other baselines for all actions. Leveraging spatial information can further improve the performance in various settings.

\vspace{-3mm}
\paragraph{Consequence Generation.}
We finally consider generating most probable consequences for each object with exception. We report the performance  in Table~\ref{table:consequence-generation}. Our model outperforms the baselines in most metrics while some of the baselines also achieve good performance. This phenomenon might be due to the uncertainty of ground truth since sometimes the consequences do not necessarily depend on the neighboring objects for every action. However, we can see that for \emph{grasp} contextual information helps.

\begin{figure*}[t!]
\begin{center}   
\resizebox{2\columnwidth}{!}{
\includegraphics[width=2\linewidth]{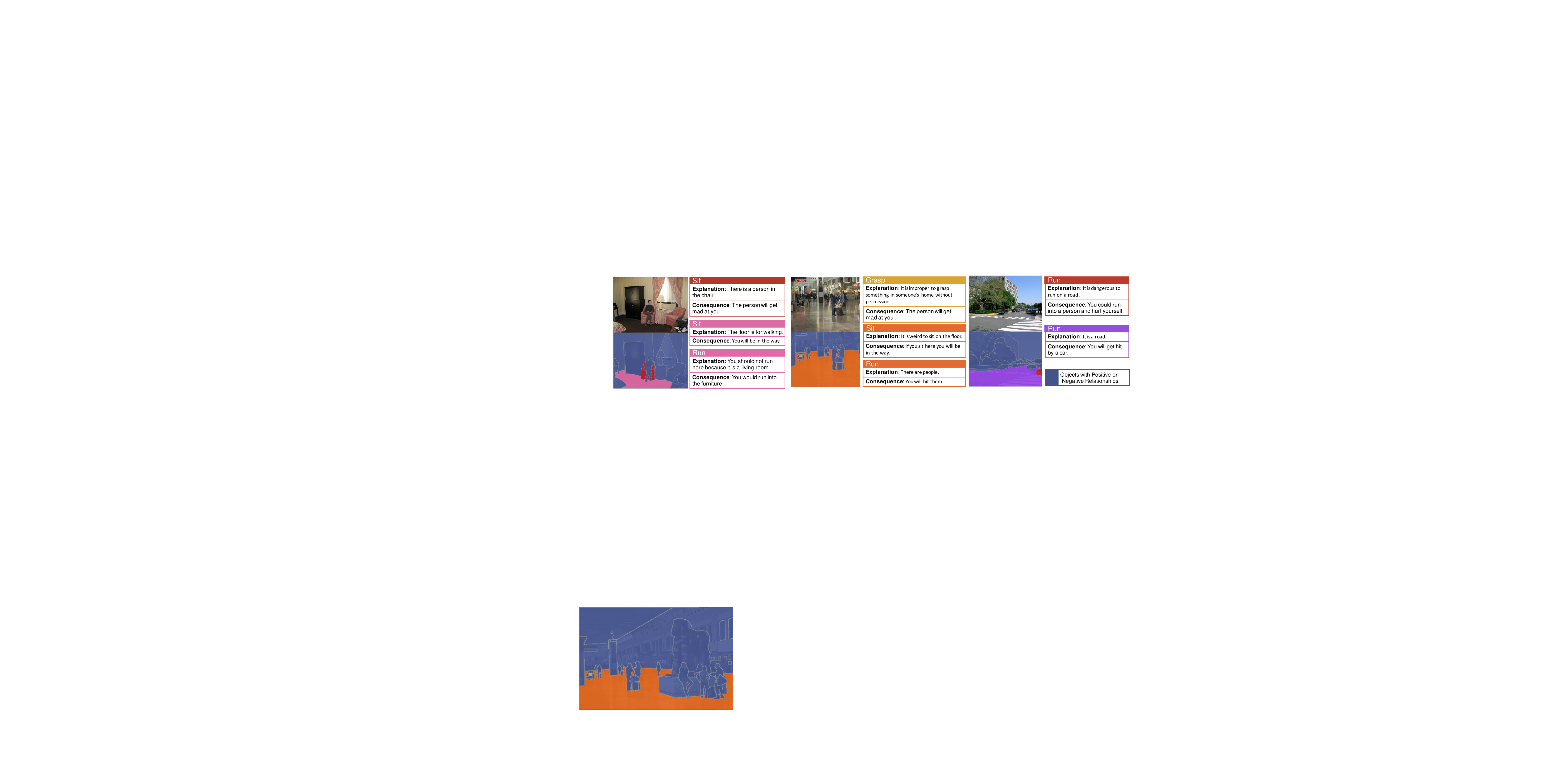}
}
\end{center}
\vspace{-3mm}
\caption{Explanation and Consequence Generation. Ground-truth segmentation is used in this experiment.}
\label{typical_exco}
\vspace{-4mm}
\end{figure*}

For both explanation and consequence generation, we found that the size of our dataset limits the diversity of the output sentences. Our model tends to describe similar situations with the same sentence. However, different from typical image captioning, we are able to generate explanations and consequences that depend on the inter-object context. Qualitative examples are shown in Fig.~\ref{typical_exco}.

\begin{table*}[t!]
\begin{center}
\centering
\resizebox{2\columnwidth}{!}{
\begin{tabular}{@{}l|cccc|cccc|cccc@{}}
%\begin{tabular}{|c|c|c|c|c|c|c|c|c|c|c|c|c|} 
\hline
 \multirow{2}{*}{Method} & \multicolumn{4}{c|}{Sit} & \multicolumn{4}{c|}{Run} & \multicolumn{4}{c}{Grasp} \\ 
  & Bleu-4 & METEOR & ROUGE & CIDEr & Bleu-4 & METEOR & ROUGE & CIDEr & Bleu-4 & METEOR & ROUGE & CIDEr\\ \hline\hline
 Unaries & 0.127 & 0.144 & 0.332 & 0.214 & 0.149 & 0.174 & 0.376 & 0.350 & 0.679 & 0.5 & 0.773 & 1.838 \\ \hline
 Chain RNN & 0.133 & 0.142 & 0.365 & 0.228 & 0.110 & 0.141 & 0.358 & 0.30 & 0.702 & 0.485 & 0.746 & 1.890\\ \hline
 FC GGNN & 0.150 & 0.150 & 0.358 & \textbf{0.231} & 0.141 & 0.164 & 0.362 & 0.309 & \textbf{0.738} & 0.528 & 0.814 & 2.046 \\ \hline
 Spatial GGNN & \textbf{0.160} & \textbf{0.155} & \textbf{0.376} & 0.220 & \textbf{0.176} & \textbf{0.181} & \textbf{0.394} & \textbf{0.409} & 0.735 & \textbf{0.532} & \textbf{0.816} & \textbf{2.069} \\ \hline
\end{tabular}
}
\end{center}
\vspace{-2.5mm}
\caption{\small Explanation Generation. In this experiment we use ground-truth instance segmentation in all models. }
\label{table:explanation-generation}
\vspace{-2mm}
\end{table*}

\begin{table*}[t!]
\begin{center}
\centering
\resizebox{2\columnwidth}{!}{
\begin{tabular}{@{}l|cccc|cccc|cccc@{}}
%\begin{tabular}{|c|c|c|c|c|c|c|c|c|c|c|c|c|} 
\hline
 \multirow{2}{*}{Method} & \multicolumn{4}{c|}{Sit} & \multicolumn{4}{c|}{Run} & \multicolumn{4}{c}{Grasp} \\
  & Bleu-4 & METEOR & ROUGE & CIDEr & Bleu-4 & METEOR & ROUGE & CIDEr & Bleu-4 & METEOR & ROUGE & CIDEr\\ \hline\hline
 Unaries & 0.130 & \textbf{0.158} & 0.375 & 0.235 & 0.167 & 0.197 & \textbf{0.427} & 0.389 & 0.492 & 0.307 & 0.528 & 0.772 \\ \hline
 Chain RNN & 0.118 & 0.150 & 0.355 & 0.183 & 0.163 & 0.156 & 0.392 & 0.340 & 0.574 & 0.342 & 0.581 & 1.023\\ \hline
 FC GGNN & 0.132 & 0.147 & \textbf{0.385} & 0.224 & 0.170 & \textbf{0.198} & 0.418 & 0.390 & 0.633 & 0.392 & 0.627 & 1.381 \\ \hline
 Spatial GGNN & \textbf{0.134} & 0.151 & 0.380 & \textbf{0.24} & \textbf{0.175} & 0.173 & 0.414 & \textbf{0.392} & \textbf{0.70} & \textbf{0.427} & \textbf{0.695} & \textbf{1.504} \\ \hline
\end{tabular}
}
\end{center}
\vspace{-2.5mm}
\caption{\small Consequence Generation. In this experiment we use ground-truth instance segmentation in all models.}
\label{table:consequence-generation}
\vspace{-2mm}
\end{table*}

\begin{table*}[t!]
\begin{center}
\centering
\resizebox{2.05\columnwidth}{!}{
\begin{tabular}{@{}l|cccccccc|cccc|cccc@{}}
%\begin{tabular}{|c|c|c|c|c|c|c|c|c|c|c|c|c|} 
\hline
 \multirow{3}{*}{Method} & \multicolumn{8}{c|}{Exception (mAcc)} & \multicolumn{4}{c|}{Explanation (CIDEr)} & \multicolumn{4}{c}{Consequence (CIDEr)} \\
  & Avg. & Avg. (E) & Sit & Sit (E) & Run & Run (E) & Grasp & Grasp (E) & Avg. & Sit & Run & Grasp & Avg. & Sit & Run & Grasp\\ \hline\hline
 independent & \textbf{0.693} & \textbf{0.408} & \textbf{0.745} & \textbf{0.461} & \textbf{0.730} & \textbf{0.452} & 0.604 & 0.31 & \textbf{0.899} & 0.220 & \textbf{0.409} & \textbf{2.070} & \textbf{0.712} & 0.240 & \textbf{0.392} & \textbf{1.504} \\ \hline
 MT & 0.676 & 0.389 & 0.714 & 0.425 & 0.697 & 0.433 & 0.616 & 0.311 & 0.604 & 0.234 & 0.401 & 1.177 & 0.397 & \textbf{0.252} & 0.343 & 0.596 \\ \hline
 MA, MT & 0.683 & 0.400 & 0.724 & 0.447 & 0.702 & 0.431 & \textbf{0.622} & \textbf{0.321} & 0.614 & \textbf{0.241} & 0.358 & 1.243 & 0.406 & 0.22 & 0.31 & 0.687\\ \hline
\end{tabular}
}
\end{center}
\vspace{-2.5mm}
\caption{\small Multi-Task Results. In this experiment we use ground-truth instance segmentation in all models.}
\label{table:multitask-result}
\vspace{-3mm}
\end{table*}

\vspace{-4mm}
\paragraph{Ablative Analysis.}
We study the effect of different choices in our model in Table~\ref{table:ablative-analysis}. In these experiments, we focus on the task of relationship prediction for \emph{sit}. We use ground-truth segmentation. Here, OC stands for object class, OR represents object's ResNet feature, and GR indicates ResNet feature extracted from the full image.

We first evaluate the performance by removing  information from the input to the model. We can see that our GGNN still successfully makes a prediction  without the global image feature while the performance of Unaries significantly decreases in this case. Unaries rely on this information more heavily, since it is its only source of context. 
As expected,  removing the class information of the object's instance (w/o OC) harms the performance most significantly. Not using the object feature (OR) as input does not seriously decrease the performance with GR, but largely harms the performance of Unaries.
%Without knowing what the class is (w/o OC) also harm the performances, point out the importances of object class predictor. Without accessing to the object feature (OR) does not seriously decrease the performance with GR, but largely harm the performance of Unaries with the absence of GR. Comparing these performances between Unaries and GGNN reveals that spatial GGNN can efficiently utilize neighbor informations to predict exception comparing to Unaries.

The number of propagation steps $T$ is an important hyper-parameter for GGNN. Table~\ref{table:ablative-analysis} shows the performance of our model with different $T$. We found that the performance saturates around $T = 3$, so we fix it as the hyper-parameter for all graph structured networks.

\begin{table}
\begin{center}
%\centering
\resizebox{0.8\columnwidth}{!}{
%\begin{tabular}{|c|c|c|}
\begin{tabular}{@{}lccc@{}} 
 \hline
 Method &  mAcc & mAcc-E \\ \hline\hline
 
 Unaries w/o OC, GR & 0.565 & 0.276 \\ \hline
 Unaries w/o OR, GR & 0.664 & 0.328 \\ \hline
 Unaries w/o OC & 0.598 & 0.306 \\ \hline
 Unaries w/o OR & 0.719 & 0.415 \\ \hline
 Unaries & 0.730 & 0.428 \\ \hline\hline
 
 Spatial GGNN, T=3 w/o OC, GR & 0.618 & 0.326 \\ \hline
 Spatial GGNN, T=3 w/o OR, GR & 0.719 & 0.396 \\ \hline
 Spatial GGNN, T=3 w/o OC & 0.622 & 0.341 \\ \hline
 Spatial GGNN, T=3 w/o OR & 0.720 & 0.424 \\ \hline
 Spatial GGNN, T=3 & \textbf{0.745} & \textbf{0.461} \\ \hline\hline
 Spatial GGNN, T=1 & 0.723 & 0.428 \\ \hline
 Spatial GGNN, T=2 & 0.723 & 0.433 \\ \hline
 Spatial GGNN, T=4 & 0.738 & 0.431 \\ \hline
\end{tabular}
}
\end{center}
\vspace{-1mm}
\caption{\small Ablative Analysis on Sit Action Relation Prediction.}
\label{table:ablative-analysis}
\vspace{-5mm}
\end{table}

\vspace{-3mm}
\paragraph{Combination with Inferred Segmentation. } We now tackle the most challenging task, i.e., going from raw images to the full visual explanation. 
%We combine our models with segmentation model to form a complete pipeline of prediction: from raw image to object level prediction. 
Since the instance segmentation in ADE dataset is quite challenging due to the limited number of images and more than 100 different classes, we use the instance segmentation model~\cite{qi2017fcis} which is trained on MS-COCO. Since  instance segmentation models only consider discrete objects, we trained an additional semantic segmentation network~\cite{yu2017dilated} on ADE20K to segment 25 additional ``stuff''  classes such as \emph{floor} and \emph{road}. To combine both results, we overlay instance segmentation results over the predicted semantic segmentation map.

We show the performance in Table~\ref{table:segmentation-relation-prediction}. Note that the instance segmentation model that we use is trained to predict 60 out of 150 ADE classes. To calculate the accuracies, we only focus on these objects. Given the difficulty of the task, the model achieves a reasonable performance. However, it also opens exciting challenges going forward in performing such a complex reasoning task.

\vspace{-4mm}
\paragraph{Multi-Task Learning.}
We now train a joint model, predicting relationships, generating explanations and consequences in a unified GGNN. We consider two different settings. The first one is single action multi-task learning (SA-MT): sharing GGNN parameters across relationships, explanation, and consequence for each action, but not across actions. The second one is multi-action multi-task learning (MA-MT): sharing GGNN parameters across three different tasks among all the actions. The performance is shown in Table~\ref{table:multitask-result}. We can see that both SA-MT and MA-MT reach performance comparable to independent models,  even surpassing independent models in some tasks. This show that it is practical to share the parameters of a single GGNN across various tasks, thus saving on inference time.

\vspace{-1mm}
\section{Conclusion}
\label{sec:conc}
\vspace{-1mm}
In this paper, we tackled the problem of action-object affordance reasoning from images. In particular, our goal was to infer which objects in the scene a chosen action can be applied to, by taking into account scene-dependent physical and implied social constraints. We collected a new dataset building on top of ADE20k~\cite{zhou12017ade} which features affordance annotation for every object, including scene-dependent exception types with detailed explanations, as well as most probable consequences that would occur if the action would be taken to an object. We proposed a model that performs such visual reasoning automatically from raw images, by exploiting neural networks defined on graphs. We provided extensive evaluation of our approach through various comparisons, pointing to challenges going forward.

\vspace{-1mm}
\section*{Acknowledgements}

This research was partially supported by the DARPA Explainable
Artificial Intelligence (XAI) program, and NSERC. We gratefully acknowledge the
support from NVIDIA for their donation of the GPUs used for this
research. We thank Relu Patrascu for infrastructure support.

{\small
\bibliographystyle{ieee}
\bibliography{egbib}
}

\end{document}